\documentclass[a4paper, 10pt, conference]{ieeeconf}      

\usepackage[utf8]{inputenc} 
\usepackage[T1]{fontenc} 

\usepackage{fancyhdr}
\fancypagestyle{withfooter}{
  
  \fancyfoot[C]{\footnotesize Accepted to the IEEE ICRA Workshop on Field Robotics 2024}
}

\usepackage{graphicx}
\usepackage[hidelinks]{hyperref}
\usepackage[super]{nth}
\usepackage[acronym]{glossaries}
\glsdisablehyper

\IEEEoverridecommandlockouts                              

\newacronym{ugv}{UGV}{Unmanned Ground Vehicle}
\newacronym{sugv}{SUGV}{Small Unmanned Ground Vehicle}
\newacronym{lugv}{LUGV}{Large Unmanned Ground Vehicle}
\newacronym{uav}{UAV}{Unmanned Aerial Vehicle}
\newacronym{suav}{SUAV}{Small Unmanned Aerial Vehicle}
\newacronym{luav}{LUAV}{Large Unmanned Aerial Vehicle}

\title{\LARGE \bf
A multi-robot system for the detection of explosive devices*
}

\author{Ken Hasselmann$^{1}$, Mario Malizia$^{1}$, Rafael Caballero$^{2}$, Fabio Polisano$^{3}$, Shashank Govindaraj$^{3}$,\\ Jakob Stigler$^{4}$, Oleksii Ilchenko$^{5}$, Milan Bajic$^6$, and Geert De Cubber$^{1}$
\thanks{*Author contributions: KH led the writing and drafted the article, MM drafted the article with KH, all authors participated in the elaboration of the presented ideas, provided comments and revised the article.}
\thanks{$^{1}$Ken Hasselmann, Mario Malizia, and Geert De Cubber are with the Robotics and autonomous systems lab from the Royal Military Academy of Belgium, Brussels, Belgium
        {\tt\small ken.hasselmann@mil.be}, {\tt\small mario.malizia@mil.be}, {\tt\small geert.de.cubber@mil.be}} %
\thanks{$^{2}$Rafael Caballero is with the Perception and AI unit in the Advanced Center for Aerospace Technologies (CATEC),
        Sevilla, Spain
        {\tt\small rcaballero@catec.aero}}%
\thanks{$^{3}$Fabio Polisano and Shashank Govindaraj are with Space Applications
Services NV/SA, Sint-Stevens-Woluwe, Belgium
        {\tt\small shashank.govindaraj@spaceapplications.com}, {\tt\small fabio.polisano@spaceapplications.com}}%
\thanks{$^{4}$Jakob Stigler is with Fraunhofer Insitute for High-Speed-Dynamics, Ernst-Mach-Institut, Freiburg, Germany
        {\tt\small Jakob.Stigler@emi.fraunhofer.de}}%
\thanks{$^{5}$Oleksii Ilchenko is with Lightnovo ApS, Birkerød, Denmark
        {\tt\small olil@lightnovo.com}}%
\thanks{$^{6}$Milan Bajic is with Scientific Council HCR—Center for Testing, Development, and Training, Zagreb, Croatia
        {\tt\small milan.bajic@ctro.hr}}%
}

\begin{document}

\maketitle
\thispagestyle{withfooter}
\pagestyle{withfooter}

\begin{abstract}

In order to clear the world of the threat posed by landmines and other explosive devices, robotic systems can play an important role. However, the development of such field robots that need to operate in hazardous conditions requires the careful consideration of multiple aspects related to the perception, mobility, and collaboration capabilities of the system.
In the framework of a European challenge, the Artificial Intelligence for Detection of Explosive Devices - eXtended (AIDEDeX) project proposes to design a heterogeneous multi-robot system with advanced sensor fusion algorithms. This system is specifically designed to detect and classify improvised explosive devices, explosive ordnances, and landmines. 
This project integrates specialised sensors, including electromagnetic induction, ground penetrating radar, X-Ray backscatter imaging, Raman spectrometers, and multimodal cameras, to achieve comprehensive threat identification and localisation. 
The proposed system comprises a fleet of unmanned ground vehicles and unmanned aerial vehicles. This article details the operational phases of the AIDEDeX system, from rapid terrain exploration using unmanned 
aerial vehicles to specialised detection and classification by unmanned ground vehicles equipped with a robotic manipulator.
Initially focusing on a centralised approach, the project will also explore the
potential of a decentralised control architecture, taking inspiration from
swarm robotics to provide a robust, adaptable, and scalable solution for explosive detection.

\end{abstract}

\section{INTRODUCTION}
\label{sec:sota}
Improvised Explosive Devices (IEDs) and Explosive Ordnances (EOs) are a major cause of casualties in conflict zones and post-conflict environments. 
Their potential to harm civilians and military personnel alike highlights the need for effective detection and clearance solutions.
In this section, we delve into robotics in hazardous environments, particularly focusing on demining applications, before reviewing some of the applications from previous decades.
\subsection{Robotics in hazardous environments}
In our context, the term "hazardous environments" refers to conditions that pose risks to human safety. These conditions include, but are not limited to, exposure to toxic substances, extreme temperatures, high radiation levels, or physical hazards.
Hazardous environments are commonly found in industrial facilities, disaster-affected areas, and regions affected by military conflicts, including landmines.
The usage and consequent development of specific robotic systems in such diverse environments have become increasingly crucial over the last decades~\cite{Trevelyan2008}, reflecting advances within the community, particularly in disaster response and routine work under challenging conditions.
\subsection{Challenges and objectives in demining robotics}
Robotics for demining faces the challenge of clearing explosive remnants, including threats like landmines and improvised explosive devices (IEDs), which pose immediate and long-term
risks and impede development long after conflicts end~\cite{Bogue2011-gc}.

The hazardous nature of counter-IED tasks involves risks ranging from immediate threats to constraints on human presence.
However, it is crucial to differentiate based on the final objective of the operation, whether it aims at rapid military clearance, ensuring safe passage for military personnel, or humanitarian demining, which aims at reaching 100\% ground clearance at an acceptable cost, as described in~\cite{university2000demining}.

The complexity associated with IEDs, including various triggers, explosive compositions, and deployment methods, contributes to the ongoing difficulties in detection.
Challenges such as varied terrains, hostile environments, and the demand for faster and more sensitive detection further accentuate the importance of advancing sensor technologies~\cite{Nicoud1997-vu}.
To address these challenges, various projects aim to provide effective solutions. In particular, the European Union has funded projects dedicated to mine action, including the TIRAMISU~\cite{tiramisu2012} project under the \nth{7} Framework Programme for Research (FP7) and the AIDED~\cite{aided2023} project funded under the Preparatory Action on Defence Research (PADR).
\subsection{Advancements in unmanned systems for demining robotics}
Initially, the mine action community explored solutions involving legged robots, focusing on teleoperation to ensure operator safety. One such example is the Dylema walking robot~\cite{GonzalezdeSantos2005}, equipped with a manipulator to scan areas using a metal detector. The project demonstrates the usage of legged robots to navigate uneven terrains based on the studies carried out by Gonzalez de Santos et al.~\cite{GonzalezdeSantos2004} the previous year.
In subsequent years, Montes et al.~\cite{Montes2015} made significant advances to their previous solution~\cite{GonzalezdeSantos2005}, improving the coordination between the scanning manipulator and the hexapod robot. This included an increased size, a reconfigured body, and improved sensor support.

In parallel, Jaradat et al.~\cite{Jaradat2012}, proposed a conceptual idea for a fully autonomous navigation robot for demining, featuring a bogie-based holonomic suspension for improved maneuverability in all directions. The design was inspired by previous wheeled platforms by Santana et al.~\cite{Santana2008} and Baudoin~\cite{Baudoin2008} that validated their stability and effectiveness in challenging environments using simulations.

Unlike previous walking prototypes, Balta et al.~\cite{Balta2014} pursued a different approach, developing Teodor, a robust robotic platform for demining tasks that leveraged 
the use of multichannel detector arrays.
In a subsequent work by De Cubber et al.~\cite{deCubber2014}, based on the existing Teodor \gls{ugv}, the use of unmanned aerial support to locate suspected mine sites was evaluated.
By involving a \gls{uav} in the setup, they successfully address the issue of a narrow field of view inherent to UGVs~\cite{Cantelli2013}, improving situational awareness and navigation strategies.

Conceptual ideas have been developed that involve the use of swarm robotics, such as the fully autonomous multi-robot
solution proposed by Mullen et al.~\cite{Mullen2011}, focusing on localising and monitoring generic targets within a specified search space.
Furthermore, Sawant et al.~\cite{Sawant2022} discussed the deployment of a swarm of robots for military use to tackle tasks that traditionally involve high causalities, such as landmine detection, highlighting the potential of robotics in hazardous environments for military applications.

Recently, the Disarmadillo~\cite{Cepolina2022} project introduces a versatile platform designed specifically for demining tasks, while also considering potential applications in agriculture.
The platform provides a modular solution capable of carrying out various tools for demining operations.
In addition, the open nature of the project encourages collaboration, adaptation of the platform and innovation.

\section{PROBLEM STATEMENT}
\label{sec:prob}

The objective of the AIDEDeX project is to design a multi-robot system for the
detection and classification of IEDs, EOs and Landmines.
As described in Section~\ref{sec:sota}, one of the main challenges in the detection of IEDs comes from the
large degree of variability in terms of their shape, size, and overall construction.
Despite the wide variations in IED construction, these devices generally share three core components: a main charge, an initiating system, and a container. The main charge can be a high explosive (for detonation and blast effects) or a low explosive (for combustion or fragmentation).
Initiating systems employ electrical, mechanical, or chemical triggers. Activation methods include remote detonation (via a command wire, radio, or cell phone), victim-operated mechanisms (pressure, tripwire, movement), or timing devices. 
Containers vary significantly, ranging from disposable casings to robust structures designed to shape blast and fragmentation patterns.
%
This variety calls for the use of a large array of sensors for potentially being able to detect metallic objects, hollow objects, camouflaged or hidden objects, chemicals, etc.
The main sensing techniques reported in the scientific literature in the context of demining are:
Electromagnetic Induction (EMI), Ground Penetrating Radar (GPR),
microwave radiometry, nuclear quadrupole resonance, odor sensors, 
electrical impedance tomography, ultrasound, X-Rays, and IR/hyperspectral cameras.
For a comprehensive explanation of the use of these different sensors,
we refer the reader to Florez and Parra~\cite{Florez2016}.

Sensor fusion techniques have the potential to significantly enhance the extraction of meaningful information from sensor data. Their application could lead to the development of more precise explosive detection and classification systems.
%
Nonetheless, significant challenges lie ahead, such as the fusion of heterogeneous data
and the potential lack of sufficiently diverse data to train machine learning models, 
which are increasingly used in sensor fusion algorithms~\cite{Alzubaidi2023,Alatise2020}.

While a single-robot system equipped with a comprehensive sensor suite is feasible, it presents several limitations. The system would likely suffer from reduced speed and area coverage due to increased payload,  demanding computational requirements for sensor data processing, and a higher risk of failure in the event of robot malfunction or damage, such as an unwanted blast.

Furthermore, in the context of military demining and counter-IED, as described in Section~\ref{sec:sota}, the system must allow rapid exploration to quickly secure a given environment.

This is why we propose a heterogeneous multi-robot system, where robots with different motor and
sensor capabilities operate in distinct phases of the counter-IED process, that could offer
advantages in speed, reliability, and processing.

By leveraging the use of fast UAVs on top of heavier ground robots, a heterogeneous multi-robot system can explore larger areas and therefore optimise the exploration scheme of the slower robots
by sharing data within the system.
Also, the system would be more reliable as the failure of one robot in the system would not comprise the whole mission. Finally, by processing sensor data onboard the robots, we distribute the computational power needed
to process data from the different sensors.

However, introducing a multi-robot system also introduces some challenges.
A key challenge inherent in multi-robot systems is the spatial and temporal correlation of sensor measurements across robots. This challenge extends to both specialised sensors and those dedicated to navigation and localisation, which are essential for creating a shared map within the system.

Another associated challenge, both for the single-robot and the multi-robot case, 
is related to navigation in hazardous and diverse environments; in fact, designing
a robotic system capable of both indoor and outdoor navigation is a challenging task~\cite{AlKhatib2020}.
Multi-robot systems offer a potential advantage in this regard, as the presence of multiple robots provides redundancy. This means that if one robot becomes obstructed or fails, others can continue exploring, increasing the robustness of the system.
One potential solution to address the precise positioning of robots and their sensors
is the use of a Real-Time Kinematic (RTK) enhanced global positioning system (GPS). This approach yields high positional accuracy but is only suitable for outdoor environments.
In case of GPS jamming or in indoor settings, a promising solution lies in the implementation of (multi-robot) Simultaneous Localisation and Mapping (SLAM) techniques~\cite{Chen2023}. These techniques facilitate concurrent mapping of the environment and robot localisation.

While a multi-robot, centrally controlled system offers advantages, a significant design limitation persists. 
Since the system depends on communication with the command centre, any disruption in connectivity could cause the system to fail.
This creates a single point of failure.

In this regard, we will also explore the design of a swarm system, that is,
an alternative decentralised control system.
In swarm robotics, the control system must be adaptive, robust, and scalable~\cite{Dorigo2021}. These properties are
due to the decentralised nature of the system.
%
One of the issues in designing robot swarms is ensuring that individual robots
must function without relying on any external infrastructure or robot, which means that
they must be individually programmed to be fully autonomous. 
The global behaviour of the swarm then emerges from the interactions
between the robots and their environment. 
This issue becomes increasingly difficult when designing heterogeneous swarms, 
such as the AIDEDeX system, where robots possess diverse capabilities and functionalities.

From all these challenges, we can extract the principal research questions
of the AIDEDeX project:

\noindent (i) How can we best leverage specialised robot capabilities to design a multi-robot system
to explore, detect, and classify IEDs and landmines, while maximising coverage and detection speed?

\noindent (ii) How can we effectively fuse heterogeneous sensor data to optimise the detection and classification of IEDs and landmines, addressing the potential for limited training data?

\noindent (iii) What navigation solutions are optimal for multi-robot operation in diverse and hazardous environments, both indoor and outdoor, with online adaptation?

\noindent (iv) How can we design a swarm system that ensures adaptability, robustness,
and mission continuation in the face of potential communication disruptions?

\section{METHODS}

The AIDEDeX project introduces a framework for the deployment of a heterogeneous fleet of unmanned vehicles, equipped with state-of-the-art sensors and sensor fusion algorithms.
It aims at detecting and classifying IEDs, EOs, and landmines. 
This multi-robot system integrates several specialised sensors such as Electromagnetic Induction (EMI), Ground Penetrating Radar (GPR), X-Ray Backscatter Imaging (XRB), Raman spectrometer, Infrared (IR), hyperspectral, and RGB Cameras, to locate and identify threats.
%

The AIDEDeX system comprises several \gls{ugv}s and \gls{uav}s. 
More precisely, one \gls{lugv}, two \gls{sugv}, one \gls{luav}, and two \gls{suav}. 

\begin{figure}[tpb]
    \centering
    \includegraphics[scale=0.16]{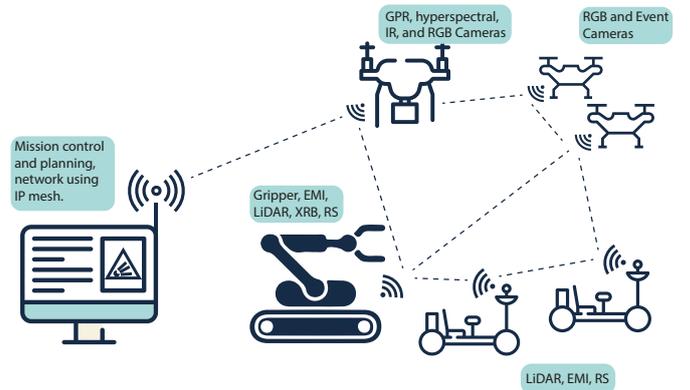}
    \caption{\textbf{High level architecture of the AIDEDeX system.} The system comprises six robots, one \gls{lugv}, two \gls{sugv}, one \gls{luav}, and two \gls{suav}. The mission control center (left) sends the mission to the multi-robot system (right). The sensors installed on the different robots are presented
    in green boxes: Electromagnetic Induction (EMI), Ground Penetrating Radar (GPR), X-Ray Backscatter Imaging (XRB), Raman Spectrometer (RS), LiDAR, Infrared (IR), hyperspectral, and RGB cameras.}
    \label{fig:arch}
\end{figure}

The general architecture of AIDEDeX will initially focus on a centralised control mechanism.
The mission planning is centrally managed through a control station, where an operator
supervises the evolution and execution of the task. Fig.~\ref{fig:arch} illustrates this high-level architecture.
All robots will be equipped with a degree of autonomy, including obstacle avoidance capabilities, and will use the Robot Operating System 2 (ROS2) for sensor and actuator integration.
One of the objectives of the project is to explore decentralised architectures.
In this case, the robots will communicate directly with each other, share sensor data, and collaboratively create a heatmap of threats in the explored area.
In the following, we will describe the different phases and subtasks of the operation and give details on the different robots that will tackle those subtasks. These descriptions are our initial design; changes may be made on the basis of the performance of our sensors achieved during field tests.

\subsection{Fast terrain exploration and first detection phase}
In order to explore the terrain that needs to be secured, the initial phase consists of a rapid exploration and a first specialised detection phase. 
For this, we will use two \gls{suav}, one \gls{luav} and two \gls{sugv}.

These \gls{suav} will reach high speed and agility to perform reconnaissance in indoor and outdoor environments.
They will be equipped with depth-sensing colour cameras and event-based cameras for navigation and
vision-based localisation. In addition, they will have an RGB camera to perform the threat search.
They will share the map of the environment and their initial vision-based threat detections with the other robots of the system.

For SUAVs, since there are no commercial solutions available that meet the specific requirements of this project, we chose to use a custom built platform by CATEC. The design will focus on maximising speed and agility while ensuring good battery life during this initial phase. Additionally, a customised design will facilitate and optimise the mounting of sensors and the on-board processing unit more efficiently. Also, by using an open-source autopilot, the system can be controlled by software developed specifically for the project.

The \gls{luav} will be equipped with a standard set of vision, navigation and localisation sensors
(including LiDAR, laser altimeter, etc.) as well as a set of specialised sensors: a 
Ground Penetrating Radar (GPR), a hyperspectral, infrared, and RGB camera.
The SUGV will also be equipped with a standard set of vision, navigation, and localisation sensors, and will
have an Electro Magnetic Interference (EMI) sensor as its specialised sensor.

\begin{figure}[tpb]
    \centering
    \includegraphics[scale=0.72]{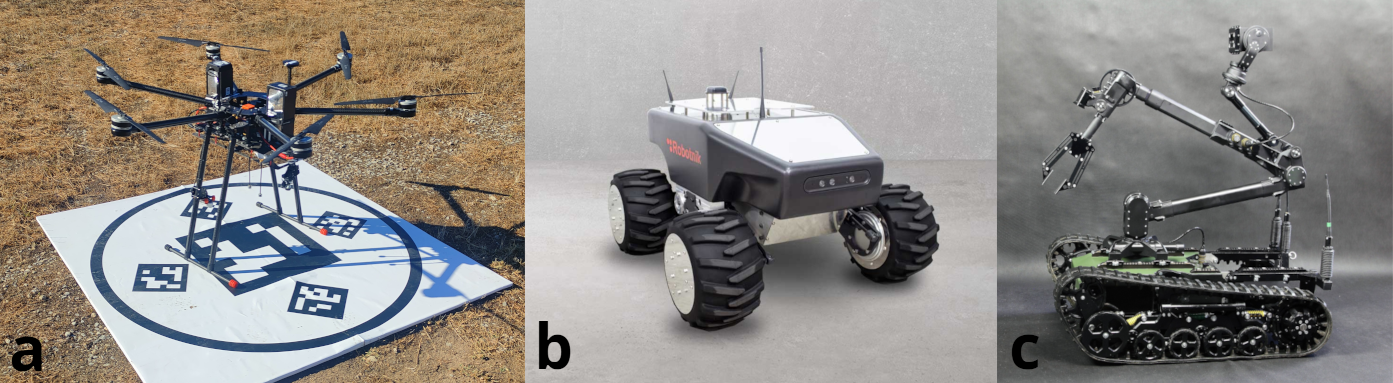}
    \caption{\textbf{The robots used in the AIDEDeX system.} \textbf{a}, the CATEC 750 \gls{uav}; \textbf{b}, the Robotnik SUMMIT XL \gls{sugv}; \textbf{c}, the PIAP Patrol \gls{lugv}. Image credits: \textbf{a}, CATEC (\url{https://www.catec.aero}); \textbf{b}, Robotnik (\url{https://robotnik.eu}); \textbf{c}, PIAP (\url{https://piap.lukasiewicz.gov.pl}).}
    \label{fig:robots}
\end{figure}

We also chose to use a custom-built platform for the \gls{luav}, 
Fig.~\ref{fig:robots}a presents an example of a custom-built \gls{uav}, the CATEC 750, built by CATEC a partner of the AIDEDeX project (see Section~\ref{sec:partners} for details on the AIDEDeX consortium).
This choice offers maximum control over the \gls{luav}, allowing the integration of the necessary sensors and onboard computers. 
Although a commercial frame may be used as a base, the custom approach ensures flexibility in payload configuration and navigation systems, which could be essential when considering indoor and GNSS-denied flight scenarios.

The \gls{luav} will scan the environment by flying over the terrain at low altitudes, based on the shared map with the \gls{suav} and detect threats using its sensors.
The \gls{sugv} will also scan the environment with its sensor based on the same map and focus on the areas with
the highest probability of threats. This probability will be computed using heuristics based on the type of environment and vision-based detections.

For the \gls{sugv}, we chose the Robotnik Summit-XL robot platform (see Fig.~\ref{fig:robots}b).
This platform offers a rugged design with a lightweight aluminium frame and has the potential to seal connections for water resistance. 
This SUGV is designed for exploration and has a battery life of approximately 5 hours. 
The Summit-XL can support a payload of 50kg, allowing one to fit the required onboard equipment.



\subsection{Second detection phase}
After the mapping and first detection phase, the objective is to confirm and possibly classify
threats. For this, we will use a \gls{lugv} with a robotic manipulator arm.
This \gls{lugv} will also be equipped with a standard set of vision, navigation, and localisation sensors, as well as
specialised sensors: X-Ray Backscatter Imaging (XRB) sensor, and a Raman spectrometer both mounted on the robotic arm and an Electro Magnetic Interference (EMI) sensor.
Based on the information provided by the other robots, the LUGV will navigate to its positions of interest based on the probability of threats. It will then use its sensor to validate and possibly classify the type of threats that were detected.

The PIAP Patrol (see Fig.~\ref{fig:robots}c) robot was selected as a potential candidate for the \gls{lugv}.
This platform offers several key advantages. Its tracked locomotion system provides versatility across various terrains (sand, gravel, clay, asphalt, and concrete).
The PIAP Patrol can also fit additional onboard equipment (computers, batteries, sensors, and communication systems).
It is rated IP65, ensuring protection against dust and rainwater. 
The expected battery life could potentially last up to half a day. 
It also comes equipped with a 6 Degrees of Freedom (DoF) robotic arm.

\subsection{Communication and autonomy}

In the initial phase, as described in Section~\ref{sec:prob}, the control of the AIDEDeX system will be centralised.
That is, during the different phases of the exploration and detection process, the robots will remain connected to the command centre using an IP mesh network to ensure maximum connectivity leveraging software defined radio with self-configuration and self-healing network with MIMO antennas network.
To ensure collaboration and task sequencing, each robot transmits its current status and sensor data to the command centre. These data are also shared between robots, enabling the command centre to coordinate phases and actions based on the overall system status.
The command centre aggregates sensor data from all robots, generating a comprehensive heatmap representing potential threats.

In the swarm approach, which we would like to reach as a final objective of the project,
one possible architecture for complete decentralisation would be to take advantage of the concept of \textit{mergeable nervous system}, introduced by Mathews et al.~\cite{Mathews2017} and subsequently implemented by Jamshidpey et al.~\cite{Jamshidpey2020}.
In this approach, robots establish a dynamic network for distributed control. This mergeable nervous system (MNS) approach allows for a degree of central coordination within a self-organised framework, combining the benefits of both centralised and decentralised multi-robot systems.

\section{TIMELINE}
\label{sec:partners}


This project is part of a technological challenge that started in December 2023.
A technological challenge is a structured research and development framework in which multiple teams address a complex objective within a shared testing environment. 
The challenge organiser pre-defines experimental protocols in collaboration with participants. 
This approach is particularly valuable for the study of complex systems, especially those integrating artificial intelligence and machine learning. It enables objective performance evaluations through key metrics, benchmarking the quality and quantity of detections using a variety of sensors.
The organization of the challenge is inspired by the structure of the DARPA Subterranean Challenge and DARPA Robotics Challenge, taking place in extreme environments.
These challenges have allowed researchers to produce valuable scientific outputs~\cite{Carlone2023} addressing the needs of the community.
Field tests will be organised each year through the duration of the project to assess the level of advancement of the four different projects that are participating in this challenge, for a total of four field tests.
%

At the end of the four years, the objective is to allow real-time detection and localisation of hidden
threats in outdoor and indoor environments. These objectives will be mitigated with
the evolution of the project based on the level of maturity that the system will be able to reach.




\section{CONCLUSIONS}

The AIDEDeX project represents a significant leap forward in the development and deployment of multi-robot systems for the detection and classification of improvised explosive devices, explosive ordnance, and landmines. 
By leveraging a heterogeneous fleet of unmanned aerial and ground vehicles equipped with advanced sensors and AI-based sensor fusion algorithms, this initiative aims to significantly improve operational efficiency, safety, and effectiveness in hazardous environments. 
The integration of various robotic platforms, each with specialised capabilities, addresses the complex challenge of detecting a wide range of explosive threats in diverse and challenging terrains.

Moreover, the exploration of both centralised and decentralised control
AIDEDeX aims to address current challenges related to communication, 
navigation, and autonomy in multi-robot systems.

Through annual field tests within the framework of a technological challenge,
the AIDEDeX project seeks to test and evaluate its system and refine its methodologies.

As the project progresses, the lessons learnt and achievements
hold promise not only for military and humanitarian demining efforts, but also for broader applications of multi-robot systems in hazardous and challenging environments.






\section*{ACKNOWLEDGMENT}
This project has received funding from the European Defence Fund (EDF) under grant agreement EDF-2022-LS-RA-CHALLENGE-DIGIT-HTDP-AIDEDex. 
Views and opinions expressed are however those of the author(s) only
and do not necessarily reflect those of the European Union or the European Defence Fund (EDF). Neither
the European Union nor the granting authority can be held responsible for them.

We wish to thank all partners of the AIDEDeX consortium, \textit{Space Applications Services}, the \textit{Royal Military Academy of Belgium}, \textit{HCR-CTRO}, \textit{CATEC}, \textit{Fraunhofer EMI}, and \textit{Lightnovo ApS}.

\center\includegraphics[scale=0.25]{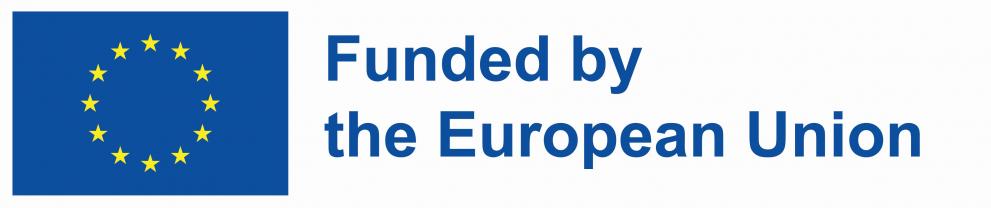}

\bibliographystyle{IEEEtran}

\bibliography{IEEEabrv, Sample}

\end{document}